# Inverse Reinforcement Learning Based Stochastic Driver Behavior Learning

Mehmet F. Ozkan*,a, Abishek J. Rocque**,a, Yao Ma***,a

*mehmet.ozkan@ttu.edu, **abishek-joseph.rocque@ttu.edu, ***yao.ma@ttu.edu
a Department of Mechanical Engineering, Texas Tech University, Lubbock, TX 79409 USA

Abstract: Drivers have unique and rich driving behaviors when operating vehicles in traffic. This paper presents a novel driver behavior learning approach that captures the uniqueness and richness of human driver behavior in realistic driving scenarios. A stochastic inverse reinforcement learning (SIRL) approach is proposed to learn a distribution of cost function, which represents the richness of the human driver behavior with a given set of driver-specific demonstrations. Evaluations are conducted on the realistic driving data collected from the 3D driver-in-the-loop driving simulation. The results show that the learned stochastic driver model is capable of expressing the richness of the human driving strategies under different realistic driving scenarios. Compared to the deterministic baseline driver behavior model, the results reveal that the proposed stochastic driver behavior model can better replicate the driver's unique and rich driving strategies in a variety of traffic conditions.

*Keywords:* Driver behavior modeling, Inverse reinforcement learning.

## 1. INTRODUCTION

Understanding driver behavior is crucial to facilitate interactive driving in mixed traffic where human-driven vehicles and autonomous vehicles interact and share the road. Autonomous vehicles should understand the intent of surrounding human drivers' behaviors to safely operate vehicles when interacting with human-driven vehicles in mixed traffic. However, the unique and rich nature of a human's internal decision-making mechanism makes it difficult to understand human driver's behaviors by autonomous vehicles. To understand such human drivers' behaviors, it is necessary to establish an accurate driver behavior model which captures the uniqueness and richness of driver behavior in real-world driving scenarios.

Over the past decades, inverse reinforcement learning (IRL) based driver behavior models have been widely used to understand human's internal-decision mechanism during the operation of vehicles (Abbeel and Ng, 2004) (Kuderer et al., 2015) (Ozkan and Ma, 2020) (Huang et al., 2020) (Sun et al., 2020). The IRL approach assumes that humans are rational planners whose goals are choosing actions to maximize their rewards or equivalent to minimize cost functions during the operation of vehicles (Ng et al., 2000). The cost function has physical meanings where it encodes the unique driving preferences of the human driver, such as preferred speed, gap distance, etc., in driving scenarios. The IRL intends to learn this cost function from drivers' demonstrated driving data and solve the optimization problem with respect to the cost function recursively to generate realistic and personalized driving behaviors. However, most of the driver behavior models with traditional IRL are deterministic and aim to acquire a single cost function with a given set of driving demonstrations. This learned single cost function fails to efficiently represent the richness of human driver behavior in practice. Some studies have been proposed to learn multiple cost functions with the IRL framework to relax the deterministic assumption in different applications (Babes et al., 2011) (Choi and Kim, 2012) (Nikolaidis et al., 2015) (Sun et al., 2020). A recent probabilistic IRL model for driver behavior modeling is proposed to learn multiple cost functions from a given set of demonstrations where each cost function is learned for a specific driver (Sun et al., 2020). This proposed model considers driver behavior's uniqueness with a learned cost function for each driver, but the rich nature of human driver behavior is not considered, and each driver is represented with a deterministic behavior model.

The objective of this study is to design a stochastic driver behavior model which captures the uniqueness and richness of human behavior in realistic longitudinal driving scenarios. The distinct contributions of this study include: 1) a stochastic inverse reinforcement learning-based driver behavior model is designed to learn a cost function distribution from driver-specific demonstrations. 2) the developed stochastic driver behavior model is applied to generate driver-specific trajectories in realistic longitudinal driving scenarios. Moreover, the performance of the proposed stochastic driver model is qualitatively analyzed and compared with a baseline deterministic driver model in the simulation studies.

In the remainder of this paper, Section 2 formulates the stochastic inverse reinforcement learning-based driver behavior model. Section 3 provides the quantitative results of the proposed learning model and the comparison study with the baseline driver behavior model. Section 4 concludes the paper.

## 2. DRIVER BEHAVIOR LEARNING

### 2.1 Vehicle Trajectory Modeling

In this study, the vehicle trajectory model is described as the longitudinal position of the vehicle and expressed as a one-dimensional quantic polynomial. The quintic polynomials have been widely applied for trajectory planning by incorporating the benefits of light computation, easy configuration, and smooth motion (Kuderer et al., 2015) (Rosbach et al., 2019) (Huang et al., 2020) (Ozkan and Ma, 2021b). The longitudinal position of the vehicle is expressed in a time interval $[t_i, t_i + T_H)$, $i = 0, 1, \ldots N\text{-}1$, for a trajectory with $N$ segments and each trajectory segment has the same length $T_H$. The vehicle longitudinal position for each trajectory segment $i$ is defined as in (1)

$$s_i(t) = \alpha_0 t^5 + \alpha_1 t^4 + \alpha_2 t^3 + \alpha_3 t^2 + \alpha_4 t + \alpha_5 \tag{1}$$

where $\alpha_{0-5}$ are the polynomial coefficients for each demonstrated trajectory segment and $t \in [t_i, t_i + T_H)$. The longitudinal velocity and acceleration can be expressed as $\dot{s}(t)$ and $\ddot{s}(t)$, respectively.

### 2.2 Stochastic Inverse Reinforcelerment Learning (SIRL) with Cost Function Distribution

In this study, the stochastic inverse reinforcement learning approach (SIRL) is proposed to learn the human driver behavior model from the demonstration. It is assumed that drivers are rational planners, and they are exponentially more likely to select trajectories with lower costs. Given a set of trajectory demonstrations $\Pi$ consists of $N$ observed trajectory segments, the objective is to learn a driver's cost function distribution where each cost function from the distribution captures the driver's behavior for each observed trajectory segment. Each cost function is defined as a linear combination of the features and their corresponding weights as in (2)

$$J_i = \theta_i^T \mathbf{f}_i(s_i) \tag{2}$$

where subscript $i$ represents the $i$th trajectory segment, $J$ is the cost function, $\theta$ is the weight vector, and $\mathbf{f}(r) = (f_1, f_2, \cdots, f_n)^T$ is the feature vector of each trajectory segment; $n$ represents the number of defined features. The goal is to find the optimal $\theta^*$, that maximizes the posterior likelihood of the demonstrations for each trajectory segment as in (3)

$$\theta_i^* = \arg\max_{\theta_i} p(\Pi_i | \theta_i) = \arg\max_{\theta_i} \prod_{k=1}^{L} p(s_i | \theta_i) \tag{3}$$

where $L$ represents the number of the planning subsegments for each demonstrated trajectory segment and each trajectory subsegment has the same length $T_P$, $p(s|\theta)$ represents the probability distribution over the trajectory segment, which is proportional to the negative exponential costs obtained along the trajectory segment based on the Maximum Entropy principle (Ziebart et al., 2008), as in (4)

$$p(s_i | \theta_i) = \exp(-\theta_i^T \mathbf{f}_i(s_i)) \tag{4}$$

The weight vector $\theta$ is usually not possible to be derived analytically, but the gradient of the optimization problem with respect to $\theta$ can be derived. The gradient can be obtained as a difference between the observed and expected feature values (Wu et al., 2020) as in (5)

$$\nabla \mathbf{f}_i = \tilde{\mathbf{f}}_i - \mathbf{f}_i^e \tag{5}$$

where $\tilde{\mathbf{f}}_i$ is the average observed feature values of the demonstrated trajectory segment $\tilde{s}_i$ as shown in (6)

$$\tilde{\mathbf{f}}_i = \frac{1}{L} \sum_{k=1}^{L} \mathbf{f}(\tilde{s}_{i,k}) \tag{6}$$

The expected feature values can be derived as the feature values of the most likely trajectory as in (7)

$$\mathbf{f}_i^e \approx \mathbf{f}_i\left(\arg\max_{s_i} p(s_i | \theta_i)\right) \tag{7}$$

The detailed derivation of the expected feature values will be later expressed in Section 2.5. The feature weight vector for each trajectory segment can be updated based on the normalized gradient descent method (NGD) (Cortes, 2006) as in (8)

$$\theta_i \leftarrow \theta_i - \eta \frac{\nabla \mathbf{f}_i}{\|\nabla \mathbf{f}_i\|} \tag{8}$$

where $\eta$ is the learning rate.

By applying the steps above, a set of $N$ different cost functions will be derived for a given set of trajectory demonstrations $\Pi$. The next step will be generating a distribution from the learned set of cost functions. To do this, it is necessary to use a suitable multivariate distribution model to fit the learned set of feature weight vectors. Copulas provide a convenient method for generating a multivariate distribution, which models the joint cumulative distribution into the marginal cumulative distributions and the copula function. Besides, copula functions can be used to describe dependency between variables that do not have the same distributions. Given that the learned feature weights do not have the same distributions and there is a dependency between the learned feature weights, the learned set of feature weight vectors $\theta = [\theta_1, \theta_2, \theta_3, ..., \theta_N]$ is then fitted using t-copula in a multivariate distribution $W_\theta$ (Bouye et al., 2000).

For the t-copula fitting, the kernel density estimation (KDE) approach is initially employed to transform the learned feature weight vectors into the copula scale [0,1] (Hill, 1985). The maximum likelihood technique is then used to fit the transformed learned feature weight vectors to the t-copula (Bouye et al., 2000). Following the fitting procedure, samples can be generated from the t-copula and then transformed back to the original scale using the inverse KDE technique.

## 2.3 Feature Construction

In this study, we focus on the car-following driving behaviors in the longitudinal direction, and therefore the following features are used to capture the relevant characteristics of the car-following driving behaviors:

**Acceleration:** The integration of the acceleration within each planning horizon is used as a feature to capture the driver's preferred acceleration and deceleration maneuvers within the trajectory subsegment.

$$f_a(t) = \int_t^{t+T_P} \|\ddot{s}(t)\|^2 dt \qquad (9)$$

**Desired Speed:** The integration of the deviation from the desired speed is used to learn the driver's preferred speed during the trip. The desired speed $v_d$ is set to the observed maximum speed of the preceding traffic in the trajectory subsegment.

$$f_{ds}(t) = \int_t^{t+T_P} \|v_d - \dot{s}(t)\|^2 dt \qquad (10)$$

**Relative Speed:** The integration of the relative speed is used to define the driver's preference for following the preceding vehicle speed $v_p$.

$$f_{rs}(t) = \int_t^{t+T_P} \|v_p(t) - \dot{s}(t)\|^2 dt \qquad (11)$$

**Steady Car-following Gap Distance:** The integration of the gap distance variation from the desired value $d_c$ is used to characterize the driver's preferred car-following gap distance $d(t)$ where $d_s$ is the minimum safety clearance and $\tau$ is the time headway. $\tau$ is defined as the observed average time headway within the trajectory subsegment.

$$d_c = \dot{s}(t)\tau + d_s \qquad (12)$$

$$f_{cd}(t) = \int_t^{t+T_P} \|d(t) - d_c\|^2 dt \qquad (13)$$

**Safe Interaction Gap Distance:** The integration of the gap distance variation from the minimum safety clearance is used to identify the driver's preferred safe interaction gap distance when the driver follows the preceding vehicle considerably closely in the congested traffic condition.

$$f_{sd}(t) = \int_t^{t+T_P} \|d(t) - d_s\|^2 dt \qquad (14)$$

**Free Motion Gap Distance:** The integration of the negative exponential growth of the gap distance is used to identify the driver's preferred gap distance to the preceding vehicle when the driver operates the vehicle in free motion rather than interacting with the preceding vehicle.

$$f_{fd}(t) = \int_t^{t+T_P} e^{-d(t)} dt \qquad (15)$$

## 2.4 Feature Selection

In this study, the trajectory segments are clustered based on the observed driving conditions, and the different sets of features are applied to each cluster. By this, three different driving conditions are considered in the longitudinal direction and the appropriate features are applied for each driving condition. The average time headway (THW) and inverse time-to-collision (TTCi) are used as main indicators to characterize the driving behavior phases for each trajectory segment. These three different driving conditions are listed and described below.

**Steady car-following:** The steady car-following phase occurs when the average THW < 6 s (Vogel, 2003) and average TTCi < 0.05 s$^{-1}$ (Lu et al., 2010) for each trajectory segment. The features $f_a$, $f_{ds}$, $f_{rs}$ and $f_{cd}$ are used to capture the driver's steady car-following behavior within the trajectory segment.

**Free motion:** In the free motion driving phase, the driver operates the vehicle without interacting with the preceding vehicle. The following conditions are used to describe the free motion driving phase within the trajectory subsegment.

1. Average THW ≥ 6 s and average TTCi ≤ 0 s$^{-1}$.
2. The average gap distance to the preceding vehicle ≥ 35 m
3. The average driver's speed ≥ 5 m/s

The first condition implies that the driver is not approaching the preceding vehicle and the driver desires larger time headway to the preceding vehicle during the trip. However, these preferred driving preferences can occur during the congested traffic situation where the driver follows the preceding vehicle closely at a lower speed with frequent stop-and-go maneuvers or in the free motion where the driver is not influenced by the preceding vehicle. To understand whether the driver is in a congested traffic situation or free motion for each trajectory segment, a K-means clustering algorithm (Lloyd, 1982) is used for a given set of demonstrated trajectory segments that satisfy the first condition. The trajectory segments are clustered in the average speed and gap distance space via the K-means algorithm. Fig. 1 shows the clustering results in the average speed and gap distance space. It can be seen that demonstrated trajectories are grouped in two clusters represented by free motion and non-free motion trajectory segments, respectively. Finally, the second and third conditions are derived by observing the K-means clustering results. The features $f_a$, $f_{ds}$ and $f_{fd}$ are used to understand the driver's free motion driving strategy for each trajectory segment.

**Unsteady car-following:** When the driver is not in steady car-following or free motion driving circumstances, the unsteady car-following phase takes place. The features $f_a$, $f_{ds}$, $f_{rs}$ and $f_{sd}$ are used to capture the driver's unsteady car-following behavior within the trajectory segment.

## 2.5 Algorithm Implementation

Before starting the learning process, each trajectory segment is partitioned into $L$ planning subsegments $[\tilde{s}_{1,1}, \tilde{s}_{1,2}, ..., \tilde{s}_{1,L}, \tilde{s}_{2,1}, \tilde{s}_{2,2}, ..., \tilde{s}_{2,L}, ..., \tilde{s}_{N,1}, \tilde{s}_{N,2}, ..., \tilde{s}_{N,L}]$ with a given collection of trajectory demonstrations $\Pi$ that consists of $N$ observed trajectory segments $(\tilde{s}_1, \tilde{s}_2, ... \tilde{s}_N)$. Algorithm 1 depicts the detailed procedures used in the driver behavior learning process.

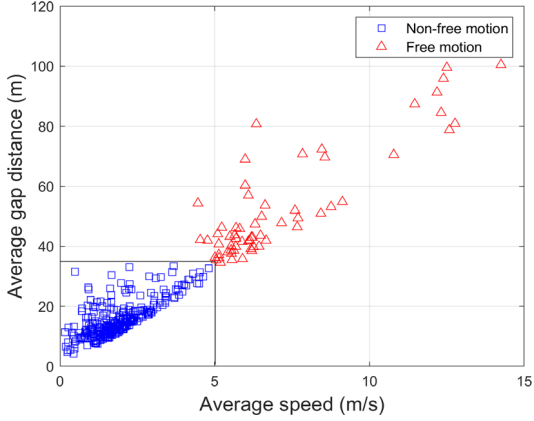

Fig. 1. The clustering results for the free motion driving.

---

Algorithm 1: Stochastic driver behavior learning algorithm

**Input:** $(\tilde{s}_{1,1}, \tilde{s}_{1,2},...,\tilde{s}_{1,L}, \tilde{s}_{2,1}, \tilde{s}_{2,2},...,\tilde{s}_{2,L},...,\tilde{s}_{N,1}, \tilde{s}_{N,2},...,\tilde{s}_{N,L})$

**Output:** $W_\theta = [W_{\theta_1}, W_{\theta_2}, W_{\theta_3}]$,
$(s_{1,1}^*, s_{1,2}^*,...,s_{1,L}^*, s_{2,1}^*, s_{2,2}^*,...,s_{2,L}^*,...,s_{N,1}^*, s_{N,2}^*,...,s_{N,L}^*)$

1: Cluster the trajectory segments with respect to the driving conditions
2: **for** each cluster **do**
3:   Initialize weight buffer $\theta \leftarrow [\ ]$
4:   **for** all trajectory segments **do**
5:     $\theta_i \leftarrow$ all-ones vector
6:     $\tilde{\mathbf{f}}_i = \frac{1}{L}\sum_{k=1}^{L} \mathbf{f}_i(\tilde{s}_{i,k})$
7:     **while** $\theta_i$ not converged **do**
8:       **for** all $s_{i,k} \in (s_{i,1}, s_{i,2},...,s_{1,L})$ **do**
9:         $(\alpha_5, \alpha_4, \alpha_3) \leftarrow$ (position, velocity, acceleration)
10:         at the initial state of the $\tilde{s}_{i,k}$
11:         Optimize $(\alpha_2, \alpha_1, \alpha_0)$ with respect to $\theta_i^T \mathbf{f}_i$
12:       **end for**
13:       $\mathbf{f}_i^e = \frac{1}{L}\sum_{k=1}^{L} \mathbf{f}_i(s_{i,k}^*)$
14:       $\Delta \mathbf{f}_i = \tilde{\mathbf{f}}_i - \mathbf{f}_i^e$
15:       $\theta_i \leftarrow \theta_i - \eta \frac{\nabla \mathbf{f}_i}{\|\nabla \mathbf{f}_i\|}$
16:     **end while**
17:     $\theta \leftarrow \theta_i$
18:   **end for**
19: **end for**
20: $W_\theta \leftarrow$ Fit each cluster's set of weight vectors $\theta$ into t-copula distribution

---

### 2.6 Trajectory Generation

In the previous section, we learned the cost function distribution that best represents the driver's preferences using the demonstrated driving data. Next, we will use the learned cost function distribution to generate driver-specific vehicle trajectories with the nonlinear model predictive control (NMPC) algorithm due to the nonlinearity introduced by the learned cost function. In the car-following scenario, it is assumed that the driver can correctly predict the motion of the preceding vehicle if the preview time horizon is reasonably small (Sadigh et al., 2016) (Ozkan and Ma, 2021a). For each trajectory segment $i$, the optimization problem needs to be solved recursively at each time instance $t$ within the prediction time horizon $T_P$ to generate the driver-specific trajectories, as shown below

$$\begin{aligned}
a_H^*(t) &= \arg\min J_H(a_H(t)) \\
J_H &= \theta_i^T \mathbf{f}_i \\
\text{subject to}&: \\
d_s &\leq d(t) \\
V_{\min} &\leq V_H(t) \leq V_{\max}
\end{aligned} \quad (16)$$

where $a_H^*(t)$ is the optimal acceleration with respect to the cost function $J_H$ from the learned cost function distribution; $d_s$ is the minimum gap distance that guarantees the safety clearance during the car-following scenario; $V_{\min}$ and $V_{\max}$ are the minimum and maximum speed constraints, respectively; $\theta_i$ and $\mathbf{f}_i$ are the random feature weight vector from the distribution $W_\theta$ and the feature vector based on the observed driving conditions within the planning horizon, respectively, as defined in Sections 2.2 and 2.4.

At the time $t+1$, the discretized inter-vehicle dynamics model with sample time $T_s$ is used for vehicle state updating as shown below

$$\begin{aligned}
d(t+1) &= (V_{PV}(t) - V_H(t))T_s + d(t) \\
V_H(t+1) &= V_H(t) + a_H^*(t)T_s
\end{aligned} \quad (17)$$

where $V_H$ and $V_{PV}$ are the longitudinal velocity of the human-driven vehicle and preceding vehicle, respectively.

### 3. RESULTS AND DISCUSSION

#### 3.1 Driver Behavior Model Implementation

In this study, we used a driver-in-the-loop simulator accomplished with MATLAB Automated Driving Toolbox to collect realistic driving data for training and testing the driver behavior model. The driver controls the vehicle in the 3D simulation environment using the hardware setup which is capable of providing a realistic driving experience, consists of a driving seat, three curved monitors, a steering wheel, and pedals, as shown in Fig. 2. The simulation cases focus on the single-lane car-following scenarios where a driver follows a preceding vehicle that is operated under several predefined speed profiles. Fig. 3 shows an example scene from the 3D simulation environment where a driver (blue vehicle) follows a preceding vehicle (red vehicle) on the road. The driving data set is collected under nine different driving scenarios, and each driving scenario is performed 30 different times by the driver. The data is collected from the simulation environment at 10 Hz. A total of 270 leader-follower trajectories are used to develop the driver model. Fig. 4 shows the speed trajectories of one of the driving scenarios used in the learning model. It can be seen that human driver behavior is quite rich in terms of driving preferences such as speeds, accelerations, etc. when the driver operates the vehicle in the same driving scenario.

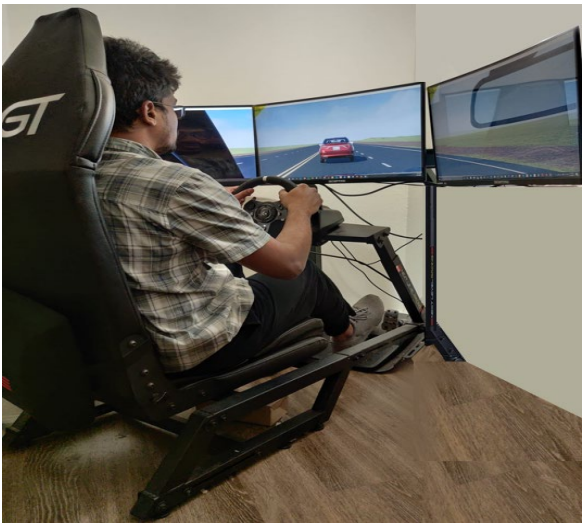

Fig. 2. A driving scene when the driver operates the vehicle in the 3D simulation environment.

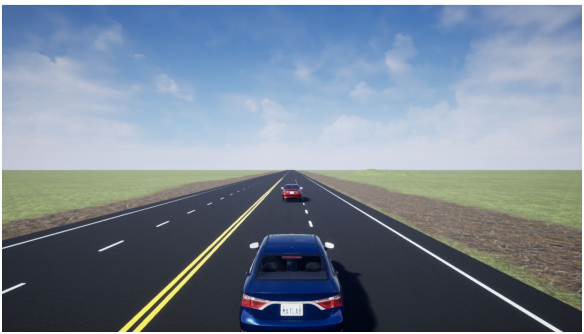

Fig. 3. A road scene from the 3D simulation environment.

To assess the learning performance of the proposed driver behavior model in various driving scenarios, 25 trajectories for each driving scenario are randomly selected for the training and the remaining 5 trajectories for each driving scenario are used for the testing. For trajectory optimization with respect to spline parameters as in step 11 of the algorithm, the BFGS Quasi-Newton method (Fletcher, 1987) is used. In the weight vector update, the learning rate $\eta$ is set to 0.2 at the initial and then drops by half for every five epochs. The length of each trajectory segment $T_H$ and subsegment $T_P$ are set to 3 seconds and 1 second, respectively; the sample time is set to $T_s$ 0.1 second and the safe distance between the vehicles $d_s$ is set to 5 m; and $V_{min}$ is set to 0 m/s and $V_{max}$ is set to the maximum speed of the preceding vehicle during the trip. For the trajectory generation, 50 samples for each driving scenario are generated by using the NMPC design, as introduced in (16).

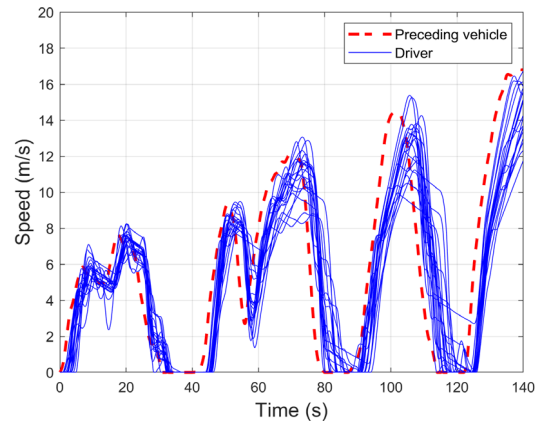

Fig. 4. A sample of driving scenarios in the data set.

3.2 Driver Behavior Model Validation and Assessment

In this section, we will evaluate the performance of the proposed SIRL based driver behavior model in various driving scenarios. Besides, we will compare our proposed SIRL based driver behavior model with the traditional IRL-based deterministic driver behavior model (DIRL), which is proposed in our previous work (Ozkan and Ma, 2020). The DIRL based baseline driver behavior model aims to learn a single cost function for a driver with a given set of demonstrated trajectory segments. In the DIRL baseline, the length of each trajectory segment is set to 3 seconds which is the same value applied to the length of the trajectory segment in the proposed SIRL based driver behavior model. For the details of the baseline driver behavior model implementation, the reader is referred to (Ozkan and Ma, 2020).

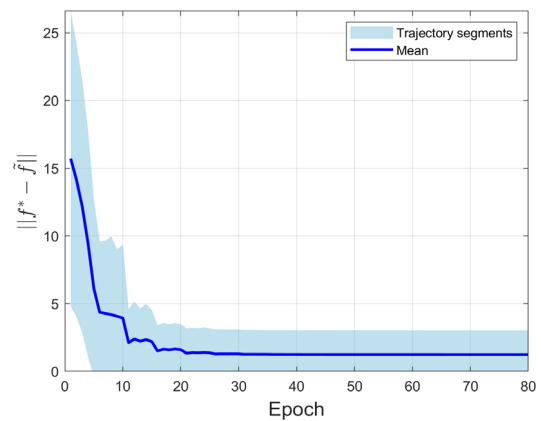

Fig. 5. Gradients ($L^2$ norm) of the trajectory segments for training.

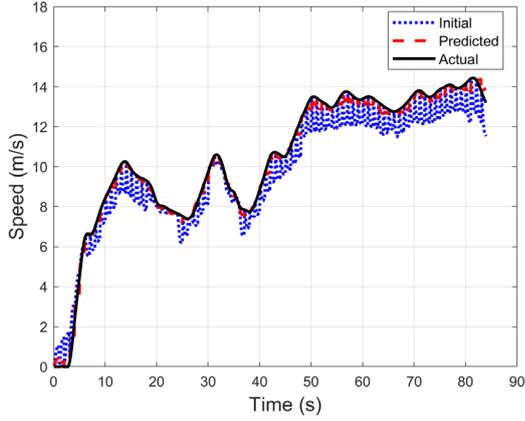

Fig. 6. Speed trajectories for training in a driving scenario.

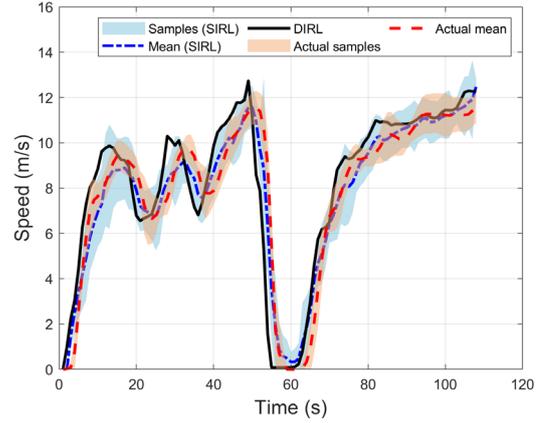

Fig. 7. Speed trajectories for testing in a driving scenario.

We first evaluate the training performance of the proposed driver behavior model. Fig. 5 shows the $L^2$ norm of weight update gradients for the trajectory segments used in training. It can be seen that the feature weights converge after roughly 30 iterations for all the trajectory segments during the optimization. Fig. 6 shows one of the driving scenarios' speed trajectories during training, where the initial guess, optimized trajectory with the learned cost function, and the actual trajectory are presented. Notably, the optimized trajectory can approach ground truth through the iterative gradients update as the weights converge to optima.

We then evaluate the testing performance of the proposed SIRL based driver behavior model with the DIRL based baseline driver behavior model in the comparison study. Fig. 7 and Fig. 8 show the actual and generated trajectories in one of the driving scenarios. The results show that the proposed SIRL based driver behavior model can generate diverse trajectory samples that cover the richness of the driver's unique driving preferences. It can be seen that the proposed SIRL based driver behavior model generates accurate trajectories against the ground truth trajectory samples even though the model has never seen the testing trajectories before, indicating a good level of validity.

To evaluate the performance of the learning model in the comparison study, Root Mean Square Error (RMSE) is used between the predicted trajectories and observed trajectories as performance matrices. Table 1 shows the average RMSE values between the mean of the observed and the predicted trajectories for the proposed SIRL model and, the average RMSE values between the predicted and mean of the observed trajectories for the DIRL based baseline model among all driving scenarios. It is found that the proposed SIRL based driver behavior model achieves 24% and 27% better speed and acceleration estimation accuracy, respectively, compared to the DIRL baseline driver behavior model. The fundamental reason is that the proposed SIRL based driver behavior model can better capture the richness of human driver behavior with the learned cost function distribution. On the other hand, the DIRL based driver behavior model fails to capture the diversity of the human driving preferences with the learned single cost function, which represents the average driving behavior.

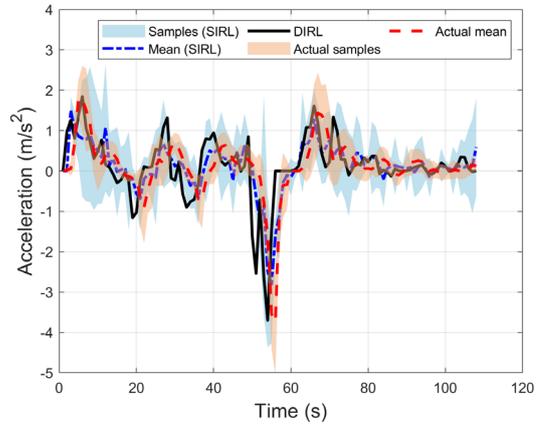

Fig. 8. Acceleration trajectories for testing in a driving scenario.

Table 1. Average RMSE values for testing in SIRL and DIRL based driver behavior models.

| Model | Speed (m/s) | Acceleration (m/s$^2$) |
|-------|-------------|------------------------|
| SIRL  | 1.43        | 0.56                   |
| DIRL  | 1.88        | 0.77                   |

Further investigating, the minor discrepancies between the predicted and demonstrated trajectories in testing for both proposed SIRL and DIRL baseline driver behavior models are observed. The prime reason for this is two-fold. First, the expected features are computed as the feature values of the most likely trajectories during the learning process. Second, the demonstrated trajectories do not guarantee to meet the optimal condition for any cost function that is a linear combination of the features that are used.

To summarize the discussion in driver behavior model assessment, the results show that the proposed stochastic driver behavior model can better learn and replicate the observed unique and rich driving preferences of a driver compared to the deterministic baseline driver behavior model in a variety of driving scenarios. The proposed driver behavior model can be beneficial for autonomous vehicles to anticipate the diverse and uncertain human drivers' behaviors in highly interactive driving scenarios.

## 4. CONCLUSION AND FUTURE WORK

In this study, a stochastic driver behavior model is proposed to learn the human driver's driving preferences in the car-following scenarios. The proposed driver behavior model uses an inverse reinforcement learning framework and obtains a cost function distribution that captures the uniqueness and richness of the human driver behavior with a given demonstrated driving data. Results show that the proposed driver model can learn and replicate human driver's unique and stochastic driving behaviors in a variety of traffic conditions. Compared with the deterministic driver behavior model, the results reveal that the proposed stochastic driver model can better learn and mimic the driving preferences of a human driver.

As a further extension of the work, a control strategy for the autonomous vehicles will be designed in interactive driving scenarios with human-driven vehicles which are modeled by the proposed stochastic driver behavior model.